# Beyond Hash Matching: AI-Based Detection of Previously Unidentified CSAM and its Legal Implications in the EU

*Emmanouela Kokolaki[1], Georgios Samaritakis and Paraskevi Fragopoulou[2]*

Foundation for Research and Technology - Hellas (FORTH), Institute of Computer Science
N. Plastira 100, Vassilika Vouton, GR-70013 Heraklion, Crete, Greece

{kokolaem, samarita, fragopou}@ics.forth.gr

## Abstract

The detection of previously unidentified Child Sexual Abuse Material (CSAM) remains a significant challenge for law enforcement authorities, online service providers, and child protection organizations, as traditional hash-matching technologies are primarily designed to identify previously verified material. This paper reviews and classifies existing AI-based technologies that could complement hash-based approaches and support the detection, analysis, and prioritization of potentially harmful content. The reviewed approaches are organized into three categories: real-time image and video analysis, image and video classification and tagging, and AI-based behavioral monitoring. A comparative analysis examines the capabilities, limitations, and potential integration of general-purpose computer vision platforms, content moderation services, and specialized CSAM detection tools within existing detection workflows.

Where technically accessible and permitted by applicable licensing and contractual conditions, an experimental evaluation was conducted using Google Cloud Vision AI and Sightengine to investigate sexual-content classification and child–adult age estimation. The experiments were conducted exclusively using datasets that did not contain CSAM. The results highlight differences in classification taxonomies and outputs across providers, as well as significant limitations in age classification, particularly for adolescents. These findings highlight the importance of using AI-based assessments as complementary tools for analysis and prioritization, rather than as definitive determinations of CSAM.

Finally, the paper examines the relevant European Union regulatory framework, with particular attention to the EU AI Act, the General Data Protection Regulation (GDPR), fundamental rights, transparency, accountability, and human oversight. The legal analysis highlights how regulatory obligations and safeguards may differ depending on the deployment context, potentially creating inconsistencies across interconnected detection and reporting workflows. Overall, the study provides a technical and regulatory overview of existing AI-based approaches and identifies opportunities and limitations for their responsible integration into CSAM detection systems.

**Keywords:** Child Sexual Abuse Material (CSAM); AI-Based Detection; Computer Vision; Content Moderation; Age Estimation; Hash Matching; EU AI Act;

---

[1] Corresponding author: kokolaem@ics.forth.gr; ORCID 0009-0003-9982-7699; Tel.: +30 281039-2033

[2] P. Fragopoulou is also with the Department of Electrical and Computer Engineering, Hellenic Mediterranean University

## 1. Introduction

The detection of previously unidentified child sexual abuse material (CSAM) has become a central objective of contemporary online safety and content moderation frameworks. Unlike the detection of known CSAM, which relies primarily on matching technologies against existing databases of verified illegal material, the identification of new CSAM requires systems capable of recognizing content that have not previously been catalogued. The importance of such detection is particularly evident in light of the fact that new CSAM frequently indicates that the sexual abuse is still ongoing, making rapid identification crucial for protecting the child involved. Therefore, mandating the detection of such material by online platforms would ensure that more victims can be swiftly safeguarded [1].

From a technical perspective, the detection of previously unknown CSAM generally relies on artificial intelligence (AI) algorithms trained to identify visual features and patterns indicative of potential child sexual abuse material and to estimate the probability that an image contains CSAM [1]. The effectiveness of these tools is closely tied to the quality of the training data on which they rely. However, the lack of publicly available CSAM datasets has significantly constrained research and development in this field, meaning that operational detection systems typically rely on proprietary datasets curated and maintained by trusted organizations. Consequently, the ongoing expansion of databases containing material confirmed as CSAM by competent authorities would substantially strengthen the capacity of automated systems to detect previously unidentified CSAM [1].

To address this challenge, technological solutions increasingly rely on artificial intelligence and machine learning techniques to identify indicators of child sexual exploitation before such material can be widely disseminated. Although the specific methodologies employed vary considerably, current approaches can broadly be grouped into three categories. First, *real-time image and video analysis systems* examine visual content as it is created, uploaded, or transmitted in order to detect potentially abusive material at the earliest possible stage. Second, *machine learning classifiers and tagging systems* assess content through predictive models that assign risk scores or probabilities based on patterns learned from previously identified CSAM. Third, *behavioral monitoring tools* focus not on the content itself but on patterns of user activity, communication, and interaction that may indicate child sexual exploitation or grooming behavior. These categories are not, however, mutually exclusive. In practice, modern detection and moderation systems may combine visual analysis, machine learning classifiers, and behavioral signals within a single detection pipeline. Together, these technologies represent the principal mechanisms through which automated systems seek to identify new CSAM, each operating at different stages of the detection process and relying on distinct forms of data analysis.

At the same time, AI-based detection of previously unidentified CSAM raises significant legal and fundamental-rights considerations, particularly where probabilistic classifications and inferences may inform content moderation, reporting, or law-enforcement action. Functionally comparable technologies may also be subject to different safeguards and ex-ante assessment requirements depending on the deploying actor and context of use. Against this background, this study combines a practical examination of selected AI-based detection and moderation services with an analysis of the legal and fundamental-rights implications of their deployment and outputs.

## 2. Methodology

The practical evaluation examined the behavior of two AI-based image-analysis and content-moderation services, Google Cloud Vision AI [2] and Sightengine [3], through API-based experiments involving two distinct scenarios: (i) sexual/nudity content classification and (ii) child–adult age classification. The evaluation aimed to investigate the classification behavior, comparability, and limitations of the selected services in tasks relevant to the identification and assessment of potentially problematic visual material. It was not intended to establish a comprehensive benchmark or overall ranking of the services, nor to determine whether particular images constituted child sexual abuse material (CSAM).

The experimental design was informed by the technical accessibility of the services, the suitability of the available datasets, and the applicable licensing and contractual conditions governing the processing of third-party images. No known or suspected CSAM was included in the evaluation.

## 2.1. Datasets and AI Services

Two publicly accessible research datasets were used, each corresponding to a separate experimental scenario. For sexual/nudity content classification, the experiment used the kshittt/sfw_nsfw_dataset [4], comprising two source subsets: Graphically Safe Images (1,102 images) and Graphically Violent Images (1,394 images). The dataset was made available on Hugging Face under a declared Creative Commons Attribution–NonCommercial 4.0 International (CC BY-NC 4.0) license and was used exclusively for non-commercial scientific research. As an additional precautionary measure, a manual image-level visual screening was conducted to identify and exclude images displaying visible indicators of potential copyright-related concerns, such as watermarks, third-party logos, or recognizable third-party copyrighted content. Following this screening, the final experimental sample comprised 288 images, with 144 selected from each of the two subsets to maintain a balanced sample. The original dataset labels were not treated as independently validated ground truth for the sexual-content categories examined.

For child–adult age classification, the FG-NET Aging Database [5] was used. The dataset contains 1,002 facial images of 82 individuals, with recorded chronological ages ranging from 0 to 69 years. All 1,002 images were included in the evaluation. The dataset contains multiple images of the same individuals at different ages and was used in accordance with its academic research conditions.

Google Cloud Vision AI was evaluated through its SafeSearch Detection functionality, particularly the adult and racy likelihood outputs. Sightengine was evaluated through its Image Moderation API for sexual/nudity content classification and its minor-probability functionality for child–adult age classification. The sexual-content experiment involved both providers, while the age-classification experiment was conducted exclusively with Sightengine.

## 2.2. Experimental Procedure and Analysis

For sexual/nudity content classification, the selected images were processed through the Google Cloud Vision AI and Sightengine APIs, and the resulting classifications and likelihood or probability scores were recorded. To enable cross-platform comparison, provider outputs were mapped to a common four-level taxonomy ranging from Safe/Non-sexual to Sexual Activity/Explicit, without assuming semantic equivalence between the underlying classifications. The analysis examined classification distributions, exact agreement, and average and maximum absolute differences between normalized levels. Of the 288 images, 284 produced valid results from both services and were included in the comparative analysis. Classification accuracy was not calculated because independently validated ground truth corresponding to the normalized taxonomy was unavailable.

For child–adult age classification, Sightengine was applied to the 1,002 images of the FG-NET Aging Database. Images depicting individuals younger than 18 years were assigned the reference class CHILD, while those depicting individuals aged 18 years or older were assigned ADULT. Predictions were obtained using Sightengine's minor-probability output and default classification decision. Four images returned no detected faces, leaving 998 valid classifications for evaluation.

Classification performance was assessed using a confusion matrix, overall accuracy, CHILD precision and recall, ADULT specificity, and CHILD F1-score. Results were additionally analyzed across five age groups (0–5, 6–12, 13–17, 18–30, and 31+ years), including mean minor probabilities, to examine age-dependent classification behavior.

### 2.3. Ethical, Licensing and Data-Governance Considerations

The evaluation used pre-existing research datasets and did not involve the collection or processing of known or suspected CSAM. The images were submitted to the APIs in their original form, without image preprocessing. Provider outputs were treated solely as analytical indicators rather than legal determinations concerning the nature or illegality of the depicted material. No images from the experimental datasets are reproduced in this article.

The applicable dataset licenses, usage restrictions, and provider-specific contractual conditions were reviewed before conducting the API-based experiments. Particular attention was given to the processing of third-party images, data retention, potential secondary use of submitted content, and the scope of the licenses required by the providers.

For Google Cloud Vision AI, the assessment considered the applicable service terms and documented data-handling practices. According to Google's service documentation, submitted images are processed in memory rather than persistently stored, and submitted content is not used to train or improve Cloud Vision models. These conditions were considered compatible with the intended non-commercial research evaluation. For Sightengine, the assessment included a review of the applicable contractual terms and written clarification obtained directly from the provider regarding the licensing requirements for processing third-party images through its API. This clarification was considered when assessing the suitability of the service for the intended non-commercial research experiments.

The kshittt/sfw_nsfw_dataset was used for non-commercial academic research under its declared CC BY-NC 4.0 license, with the additional image-level screening described above. The FG-NET Aging Database was used in accordance with its academic research conditions, including restrictions on redistribution. The assessment distinguished the datasets' declared licensing conditions from the requirements governing their processing through external services. The selection of the evaluated services also reflected the availability of sufficient information concerning their contractual and data-handling conditions. Services for which the necessary clarification had not been obtained were excluded from the final API-based experiments.

Overall, the evaluation incorporated dataset licensing, provider contractual requirements, and data-governance considerations into the experimental design, while recognizing that automated classifications should be interpreted as technical indicators rather than conclusive determinations of age, sexual content, or illegality.

## 3. AI-Based Approaches and Technologies

AI-based approaches for detecting previously unknown CSAM can be broadly classified into three categories according to their primary analytical focus: real-time image and video analysis, image/video classification and tagging, and behavioral monitoring. Although these approaches differ in their underlying techniques, input data and operational objectives, they are not mutually exclusive and may be combined within broader detection workflows. The following subsections examine their main characteristics, capabilities and limitations, together with representative technological solutions.

## 3.1 Taxonomy of AI-Based Detection Approaches

### *Real-time Image and Video Analysis*

Real-time image and video analysis is a major approach for detecting previously unknown CSAM. It enables visual content to be processed as it is created, transmitted, or uploaded, allowing potentially harmful material to be identified and acted upon before widespread dissemination. Unlike traditional batch-processing systems, which analyze stored data retrospectively, real-time systems operate on continuous data streams and can support immediate flagging, blocking, or escalation to human reviewers [6].

Technically, real-time visual analysis relies on computer vision and deep-learning techniques. Video streams may be decomposed into individual frames and analyzed sequentially or selectively using feature extraction, object detection, and models capable of capturing spatial and temporal information [7,8]. Recent approaches increasingly employ transformer-based architectures and Vision-Language Models (VLMs), which extend conventional CNN-based methods through global visual reasoning, semantic understanding, open-vocabulary recognition, and zero-shot classification. Their outputs are generally probabilistic assessments or confidence scores that can support automated moderation and prioritization [9].

Several commercial systems illustrate these capabilities. Clarifai provides cloud-based computer vision, natural language processing, and multimodal AI for images, videos, text, and audio. Its capabilities include image classification, object detection and segmentation, similarity analysis, OCR, and video analysis, with support for both pre-trained and custom models. In content-moderation workflows, these functions can provide semantic information and confidence scores for previously unseen content and support the prioritization of potentially harmful material. Clarifai is nevertheless a general-purpose AI platform rather than a dedicated CSAM detection system and cannot independently determine whether material legally constitutes CSAM [10].

Amazon Rekognition provides cloud-based image and video analysis, including object and scene detection, facial analysis, OCR, video analysis, and automated content moderation. Its Custom Labels functionality also enables application-specific models. Moderation outputs include confidence scores that can be used to define operational thresholds and prioritize potentially inappropriate content for further review [11].

Salus provides a CSAM-specific example of client-side real-time analysis. Developed by SafeToNet in collaboration with the Internet Watch Foundation (IWF), it uses machine-learning-based image recognition locally within the user's browser and can block suspected CSAM before it is displayed. Its client-side architecture also reduces the need to transmit sensitive visual content to cloud services. Automated classifications nevertheless remain indicators requiring appropriate review rather than independent legal determinations [12].

At the model level, single-stage object detectors such as YOLO (You Only Look Once) formulate detection as an end-to-end regression problem, directly predicting bounding boxes and class probabilities in a single forward pass. By eliminating separate region-proposal stages, YOLO enables efficient real-time inference and has therefore been widely applied to image and video analysis [13]. Transformer-based detectors such as DETR subsequently introduced end-to-end set prediction using transformer encoder-decoder architectures, while later real-time variants demonstrate the increasing applicability of transformer-based detection to latency-sensitive visual analysis.

Google Cloud Vision AI provides another general-purpose example, with capabilities including image labelling, object localization, face and text detection, web-based image analysis, and sensitive-content classification. Its SafeSearch functionality assigns likelihood estimates to predefined categories including adult, racy, violent, and medical content, providing a mechanism for preliminary content assessment and prioritization. Face detection additionally identifies facial regions and associated attributes, while OCR, object detection, and semantic labelling can provide further structured information for downstream analysis.

These capabilities can contribute to automated moderation and triage workflows, but Google Cloud Vision AI is not specifically designed to determine whether particular material constitutes CSAM. Its outputs should therefore be regarded as supplementary analytical information requiring contextual and expert assessment.

### *Image/Video Classifiers and Taggers*

Image and video classifiers extend detection beyond previously known material by analyzing visual content and assigning categories, labels, or risk scores learned from training data [14]. Unlike exact or perceptual hash matching, these systems can analyze previously unseen material for visual patterns associated with potentially harmful content.

A prominent specialized example is Thorn Safer Predict, which uses machine learning and predictive analytics to identify previously unseen suspected CSAM and other child sexual exploitation risks. Unlike Safer Match, which relies on matching techniques for known material [15], Safer Predict applies AI-based classification to unseen visual content and produces probabilistic risk assessments that can support automated triage and prioritization. The system uses curated datasets, including sources such as the National Center for Missing and Exploited Children (NCMEC), to support its predictive capabilities [16]. Its assessments nevertheless do not independently determine whether an image constitutes CSAM.

Google Content Safety API provides another specialized example. Available to approved organizations involved in combating child sexual abuse, it applies AI-based image classification to previously unseen material and generates confidence-based assessments of the likelihood that an image may depict CSAM [17]. The service is designed to complement hash-matching technologies by prioritizing potentially high-risk material for subsequent human review.

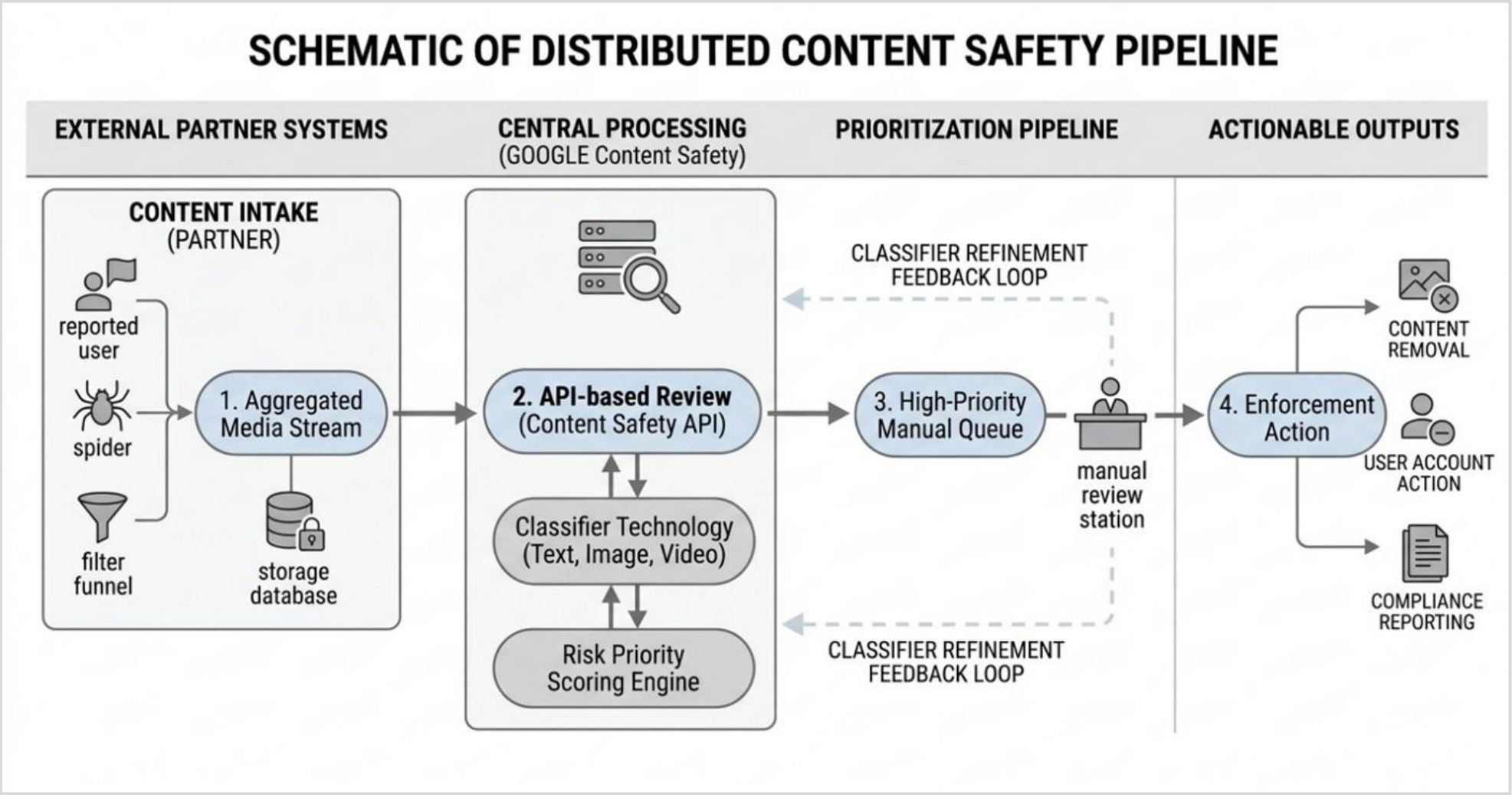

**Figure 1**. Google Content Safety API workflow, illustrating automated classifier-based prioritization followed by manual review and subsequent action by the partner. Authors' representation based on Google's description of the Content Safety API workflow [17].

AI-based image tagging provides a related but broader capability. Tagging systems generate multiple semantic labels describing objects, scenes, attributes, activities, and contextual characteristics [18]. Recent vision foundation models have extended this capability through open-vocabulary recognition. The Recognize Anything Model (RAM), for example, uses semantically enriched textual label queries to provide broader visual representations [19], which can subsequently support downstream visual analysis [20].

Microsoft Azure AI Vision provides a commercially available example of general-purpose semantic analysis. Its models can generate descriptive tags and captions, detect objects, and extract embedded text through OCR. These outputs provide contextual metadata that can support subsequent filtering, prioritization, and analysis.

Azure AI Content Safety complements these general-purpose capabilities by classifying potentially harmful visual content according to predefined categories and severity levels. Its outputs can support configurable moderation thresholds and automated filtering, although the service is not specifically designed to determine whether material constitutes CSAM.

Vision-Language Models (VLMs), including CLIP, SigLIP, and Florence-2, further extend visual analysis by jointly modelling images and natural-language concepts. They support capabilities such as zero-shot classification, open-vocabulary recognition, and richer semantic interpretation of previously unseen material. Multimodal moderation can additionally combine visual information with OCR, captions, filenames, surrounding text, or other contextual signals, potentially providing a more context-sensitive assessment than image-only classification [21].

The HIVE AI provides another example of AI-based image and video moderation. Its models classify content across predefined safety-related categories and return structured predictions with confidence scores. Additional capabilities include object and face detection, OCR, and demographic attribute classification. Face analysis may provide age-related categories and associated confidence scores, offering an additional signal for moderation and prioritization.

Sightengine is a further commercially available platform designed for automated content moderation and trust-and-safety applications. Its models analyze images and videos across categories including nudity, sexual activity, suggestive content, and violence, while complementary functions include age-related analysis, face and object detection, and OCR. The resulting classifications and confidence scores can support automated triage and prioritization, but they do not independently establish whether material constitutes CSAM.

Current AI-based CSAM detection approaches are generally better understood as multi-stage pipelines combining complementary techniques rather than relying on a single classifier. Such pipelines may combine perceptual hash matching for known material with image classification, body-part or pose detection, age estimation, and subsequent decision-making modules [22]. Age estimation and anatomical or pose analysis can provide additional visual signals, but neither can independently establish that visual material constitutes CSAM.

### *AI-based behavioral monitoring tools*

AI-based behavioral monitoring approaches complement content-based detection by analyzing network traffic, metadata, account relationships, and interaction patterns to identify anomalous activity. Techniques including Network Traffic Analysis (NTA), Deep Packet Inspection (DPI), and machine-learning-based anomaly detection can establish behavioral baselines and identify deviations potentially associated with suspicious activity [22].

Network Traffic Analysis examines communication metadata, traffic flows, connection frequencies, and temporal characteristics to identify deviations from expected behavioral patterns. Deep Packet Inspection extends analysis to packet contents and application-level characteristics and can support protocol identification, traffic classification, intrusion detection, and anomaly detection [23].

The effectiveness of payload inspection is increasingly constrained by encryption. HTTPS limits intermediary access to traffic contents, while end-to-end encryption can prevent intermediaries from inspecting communication payloads. Consequently, behavioral analysis in encrypted environments increasingly relies on observable non-content information such as traffic metadata, communication patterns, timing, and interaction characteristics.

Machine-learning-based anomaly detection identifies atypical network activity by learning patterns of normal behavior and detecting deviations that may indicate malicious activity, security incidents, or previously unknown attacks [24]. These systems can generate behavioral indicators and risk assessments from large volumes of network data.

The deployment of such techniques can raise significant data-protection concerns. Where network activity is linked to identifiable users or accounts and is used to evaluate or predict behavioral characteristics, processing may constitute profiling under Article 4(4) GDPR. The establishment of behavioral baselines, generation of risk indicators, and automated assessment of user activity therefore require careful consideration of lawfulness, necessity, proportionality, transparency, and fundamental-rights safeguards.

An illustrative example is Meta's intent-classification framework, which seeks to understand the apparent motivations of users sharing CSAM-related content. Meta has described behavioral and contextual signals including child-sexualizing search terms, attempts to obfuscate identity, and connections with groups, pages, or accounts associated with the sexualization of children. These signals are used to classify accounts according to apparent intent [25]. The precise technical implementation is not publicly disclosed.

Unlike the content-analysis technologies discussed above, behavioral monitoring is not considered here as a proposed component of the PreventCSA@EU workflow [26]. Its analysis of users or accounts, behavioral baselines, and inferred risk indicators raises distinct privacy and fundamental-rights considerations and therefore requires separate assessment under the applicable legal framework. It is consequently examined as part of the broader technological landscape rather than as a proposed extension of the PreventCSA@EU infrastructure.

### 3.2. Comparative Analysis

The analyzed platforms differ in purpose, deployment model, analytical capabilities, and relevance to CSAM detection. General-purpose computer-vision services provide broad capabilities such as object detection, OCR, semantic annotation, and content moderation, whereas specialized systems focus more directly on previously unknown CSAM. Table 1 summarizes the principal characteristics.

| Feature | Google Vision AI | Amazon Rekognition | Microsoft Azure AI Vision & Content Safety | Clarifai | The HIVE AI | Thorn Safer Predict | Google Content Safety API | Salus | Sightengine |
|---|---|---|---|---|---|---|---|---|---|
| **Primary Purpose** | General computer vision | General computer vision | General computer vision & content moderation | Multimodal AI platform | Content moderation | CSAM detection | CSAM detection | Browser protectio n | Content moderation |
| **AI-based** | ✓ | ✓ | ✓ | ✓ | ✓ | ✓ | ✓ | ✓ | ✓ |
| **CSAM-specific** | No | No | No | No | No | Yes | Yes | Yes | No |
| **Image Analysis** | ✓ | ✓ | ✓ | ✓ | ✓ | ✓ | ✓ | ✓ | ✓ |
| **Video Analysis** | Limited | ✓ | Limited (Video Indexer) | ✓ | ✓ | Not publicly documented | Not publicly documented | Limited | ✓ |
| **OCR** | ✓ | ✓ | ✓ | ✓ | ✓ | No | No | No | ✓ |
| **Object Detection** | ✓ | ✓ | ✓ | ✓ | ✓ | No | Not publicly documented | No | Limited |
| **Content Moderation** | SafeSearch | Moderation API | Azure AI Content Safety | Custom models | ✓ | ✓ | ✓ | ✓ | ✓ |
| **Risk / Confidence Scores** | ✓ | ✓ | ✓ | ✓ | ✓ | ✓ | ✓ | ✓ | ✓ |
| **Custom Model Training** | No | ✓ (Custom Labels) | Limited (Azure AI Custom Vision) | ✓ | Limited | Not publicly documented | No | No | No |
| **REST API** | ✓ | ✓ | ✓ | ✓ | ✓ | ✓ | ✓ | No public API | ✓ |
| **Cloud Deployment** | ✓ | ✓ | ✓ | ✓ | ✓ | ✓ | ✓ | No | ✓ |
| **Client-side Processing** | No | No | No | No | No | No | No | ✓ | No |
| **Human Review Required** | ✓ | ✓ | ✓ | ✓ | ✓ | ✓ | ✓ | ✓ | ✓ |

**Table 1**. Comparative Overview of the Analyzed AI-Based Image Analysis and Moderation Systems.

The platforms can be grouped into three broad categories according to their primary function. General-purpose computer-vision platforms include Google Cloud Vision AI [2], Amazon Rekognition [11], Microsoft Azure AI Vision [27] and Azure AI Content Safety, and Clarifai [10]. They provide broad visual-analysis capabilities beyond CSAM detection, including object detection, OCR, semantic annotation, and moderation. Their outputs therefore require integration with additional detection mechanisms and expert human assessment.

Content-moderation platforms include HIVE AI [28], Sightengine [3], and Salus [12]. These systems focus more directly on harmful or policy-violating visual content. Salus differs through its privacy-oriented client-side deployment, which enables preventive analysis without the same dependence on cloud processing.

Specialized CSAM detection systems include Thorn Safer Predict [15] and Google Content Safety API [29]. These systems are explicitly designed to assist approved organizations in identifying previously unknown suspected CSAM and prioritizing material for subsequent human examination.

Several common characteristics emerge. The examined systems generally generate probabilistic assessments rather than deterministic decisions, and their outputs do not independently establish whether particular material legally constitutes CSAM. Confidence scores and risk-based prioritization are therefore important components of their potential operational use.

The platforms nevertheless differ considerably in analytical scope. General-purpose services provide broader semantic and computer-vision capabilities, while specialized systems focus more directly on suspected CSAM. Privacy-preserving deployment models such as Salus demonstrate an alternative approach that can reduce the need to transmit sensitive content to cloud services.

Taken together, these technologies are best understood as complementary components within broader CSAM detection workflows. Hash-based methods can identify previously known material, while AI-based classifiers, moderation systems, semantic analysis, and other analytical techniques can provide additional information concerning previously unknown content. Their integration can therefore extend the analytical capabilities of existing systems while preserving expert human review as an essential component of assessment.

### 3.3 Potential Use with Hash-Checking Systems

The evaluated AI-based tools could be integrated into hash-checking ecosystem workflow, such as the PreventCSA@EU system [26], as complementary analysis components following hash-based verification. When submitted material does not produce a match against known-content hash databases, selected images or videos could be analyzed using cloud- or API-based services such as Google Vision AI, Amazon Rekognition, Microsoft Azure AI Vision and Content Safety, Clarifai, The HIVE AI, Thorn Safer Predict, Google Content Safety API, and Sightengine. Depending on the capabilities of each service, the resulting outputs may include moderation or classification labels, confidence or risk scores, object and face detections, OCR results, semantic descriptions, and other metadata. These outputs could enrich the information available to investigators and support automated triage by helping prioritize previously unknown or potentially suspicious material for further examination, while retaining human review as an essential part of the decision-making process.

Where legally authorized datasets are available, some platforms may additionally support the development of domain-specific models or classifiers tailored to investigative requirements. Salus represents a complementary but architecturally distinct approach. Rather than relying on cloud-based analysis, it performs AI-assisted content analysis directly on the client device, enabling real-time intervention before potentially harmful material is viewed while minimizing the transmission of sensitive content. Although its primary purpose differs from the investigative workflow of hash-checking systems, its client-side and privacy-preserving architecture illustrates how AI-based detection could also support preventive mechanisms within the wider ecosystem. The two architectural paths therefore demonstrate how AI-based analysis can complement existing hash-based detection capabilities through either external AI services or privacy-preserving client-side processing.

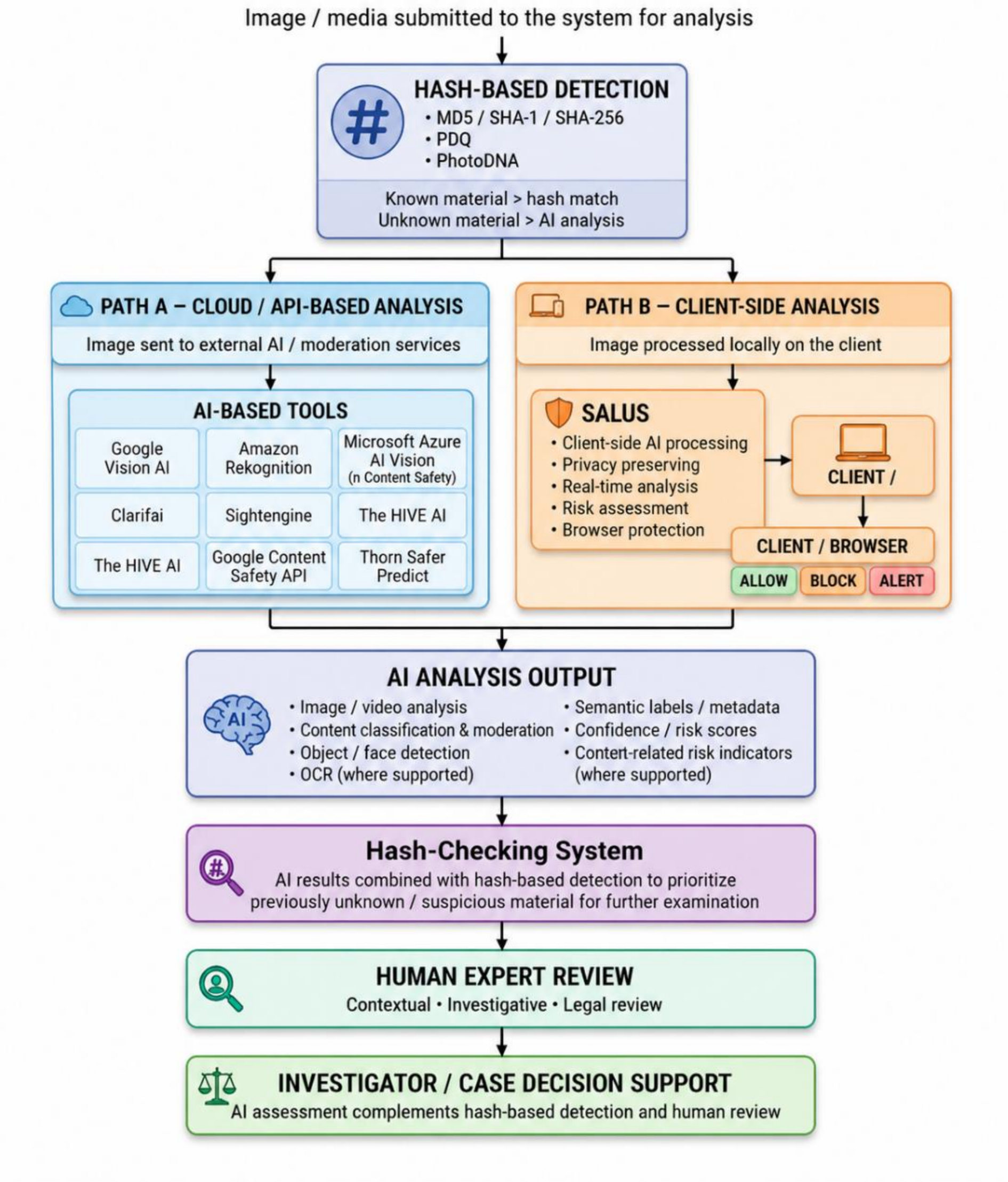

**Figure 2**. Potential integration of AI-based analysis with hash-checking detection workflow. Source: Authors' own elaboration.

## 4. Practical Evaluation: API-Based Experiments

In addition to the preceding technical analysis, a practical evaluation was conducted to examine the behavior of selected AI-based image-analysis and content-moderation platforms. The objective was not to establish a formal benchmark or an overall accuracy ranking, but to investigate how the outputs of publicly accessible AI services could support the analysis and prioritization of potentially problematic visual material.

Functional API access was not available for all examined tools. Some specialized CSAM detection solutions are restricted to approved organizations, while others require commercial accounts, controlled onboarding, or access to environments that were not available for experimentation. Consequently, the practical evaluation focused on Google Cloud Vision AI and Sightengine for sexual/nudity content classification, and Sightengine for child-adult age classification.

Two scenarios were investigated through API-based experiments using larger datasets. The first examined sexual/nudity content classification and the degree of agreement between two moderation platforms. The second evaluated the classification of persons as children or adults using an age-labelled facial image dataset. Together, these experiments investigated the practical behavior of the services, the comparability of their outputs, and their potential limitations in applications related to CSAM detection.

### 4.1. Scenario 1 - Detection and Classification of Sexual/Nudity Content

This scenario evaluated the ability of Google Cloud Vision AI and Sightengine to identify and classify sexual, explicit, nudity-related, and suggestive visual content. The objective was to examine the outputs produced by the two services and assess the degree of agreement between them, rather than to establish an absolute accuracy ranking.

Since the platforms use different classification schemes and levels of granularity, a common analytical framework was required to facilitate comparison.

#### *Normalized Classification Taxonomy*

Google Cloud Vision AI and Sightengine differ in the terminology and granularity of their content-moderation outputs. Google Cloud Vision AI provides SafeSearch classifications such as adult and racy, with likelihood values ranging from Very Unlikely to Very Likely. Sightengine offers specialized categories covering nudity, sexual activity, suggestive content, and related classes.

To facilitate cross-platform comparison, a common four-level taxonomy was introduced. The levels represent an operational normalization of the providers' outputs, rather than an assertion that the underlying models perform equivalent classifications.

| **Normalized Level** | **Normalized category** | **Description** |
|---|---|---|
| **1** | **Safe / Non-sexual** | No meaningful sexual or nudity-related content |
| **2** | **Suggestive/Partial Nudity** | Lingerie, underwear, bikini, cleavage, shirtless male, suggestive pose, or similar content |
| **3** | **Nudity/Sexual Display** | Exposed breasts, buttocks, genitalia, or a clearly undressed person |
| **4** | **Sexual Activity/ Explicit** | Sexual acts, intercourse, masturbation, or other explicitly sexual activity |

**Table 2**. Normalized Taxonomy for Cross-Platform Comparison of Content Moderation Outputs.

The taxonomy was used exclusively as an analytical layer. The original vendor classifications and their associated confidence or likelihood values were retained separately. Consequently, the normalized levels should be interpreted as an operational mapping of provider outputs, not as evidence of equivalence between the underlying models.

### *Google Cloud Vision AI normalization*

Google SafeSearch adult and racy likelihoods were used to derive the normalized level. SafeSearch does not provide a dedicated sexual-activity classifier or the fine-grained distinctions available from some other platforms. The mapping therefore reflects the strongest interpretation supported by the available signals.

| Google SafeSearch Adult | Google SafeSearch Racy | Normalized Level | Normalized Category |
|---|---|---|---|
| VERY_UNLIKELY or UNLIKELY | VERY_UNLIKELY or UNLIKELY | **1** | Safe / Non-sexual |
| POSSIBLE | VERY_UNLIKELY, UNLIKELY, or POSSIBLE | **2** | Suggestive / Partial Nudity |
| LIKELY | VERY_UNLIKELY, UNLIKELY, POSSIBLE, or LIKELY | **3** | Nudity / Sexual Display |
| VERY_LIKELY | VERY_UNLIKELY, UNLIKELY, POSSIBLE, or LIKELY | **3** | Nudity / Sexual Display |
| VERY_LIKELY | VERY_LIKELY | **4** | Sexual Activity / Explicit |

**Table 3**. Mapping of Google SafeSearch Adult and Racy Likelihoods to the Normalized Content Moderation Levels.

The Level 4 assignment should not be interpreted as Google explicitly identifying sexual activity. It represents the strongest normalized category derivable from the SafeSearch signals. Raw Google likelihoods were retained to preserve this distinction.

Google's spoof likelihood was not used to automatically reduce a sexual-content classification. Although it may provide contextual information, it is not a dedicated art/non-photographic classifier; using it to alter the sexual-content level would therefore introduce an interpretation not directly supported by the provider's output.

### *Sightengine normalization*

Sightengine provides a more granular sexual-content taxonomy. Sexual activity was mapped to Level 4, while sexual display and erotica were mapped to Level 3. Very suggestive, suggestive, and mildly suggestive content were mapped to Level 2. The absence of relevant sexual or nudity-related signals was assigned Level 1.

Specific categories such as lingerie, male underwear, cleavage, bikini, male chest, and suggestive pose were retained as secondary attributes. The nudity_art category was similarly retained as contextual information rather than automatically assigning Level 3 or Level 4. Strong artistic-nudity signals without strong sexual-display or explicit-content signals were therefore treated as contextual rather than independently sexual.

### *Dataset and experimental procedure*

The experiment used the kshittt/sfw_nsfw_dataset [4], comprising Graphically Safe Images (1,102 images) and Graphically Violent Images (1,394 images). The dataset was made available on Hugging Face under a declared CC BY-NC 4.0 license and was used exclusively for non-commercial scientific research.

As part of the data preprocessing procedure, an image-level visual screening was conducted to identify and exclude images displaying visible indicators of potential copyright-related concerns, such as watermarks, third-party logos, or recognizable copyrighted content. Following the application of these exclusion criteria, the final sample comprised 288 images, with 144 selected from each of the two subsets to maintain a balanced sample.

All 288 images were evaluated by Google Cloud Vision AI. Four images could not be processed successfully by Sightengine due to format or size limitations, leaving 284 images with results from both platforms. The normalized classifications were then compared using the common four-level taxonomy described above.

| Platform | Safe (Level 1) | Suggestive (Level 2) | Nudity (Level 3) | Explicit (Level 4) | Total Evaluated |
|---|---|---|---|---|---|
| **Sightengine** | 142 | 22 | 21 | 99 | 284 |
| **Google Vision** | 122 | 38 | 6 | 122 | 288 |

**Table 4**. Distribution of Normalized Content Moderation Levels Across the Two Platforms.

The distributions reveal differences in how the two platforms classified the images. Sightengine assigned 142 images to Level 1 and 99 to Level 4, whereas Google Vision assigned 122 images to each of these levels. Google also produced more Level 2 classifications (38 compared with 22), while Sightengine assigned more images to Level 3 (21 compared with 6).

For direct comparison, only the 284 images for which both services produced results were included in the agreement analysis.

| Comparison | Exact agreement | Agreement | Average absolute level difference | Maximum difference |
|---|---|---|---|---|
| Sightengine-Google | 231/284 | **81.3%** | **0.20** | 2 |

**Table 5**. Agreement Between Sightengine and Google Vision on Normalized Levels

The two platforms assigned the same normalized level to 231 of the 284 jointly evaluated images, corresponding to an exact agreement rate of 81.3%. In the remaining 53 cases (18.7%), their classifications differed. The average absolute difference between their normalized levels was 0.20, while the maximum observed difference was two levels.

Google produced more Level 2 classifications and fewer Level 3 classifications than Sightengine. However, these differences should not be interpreted as evidence that either platform is inherently more sensitive or stricter. They may reflect differences in the platforms' underlying classification taxonomies, confidence thresholds, and the mapping required to place their outputs within the common four-level scheme.

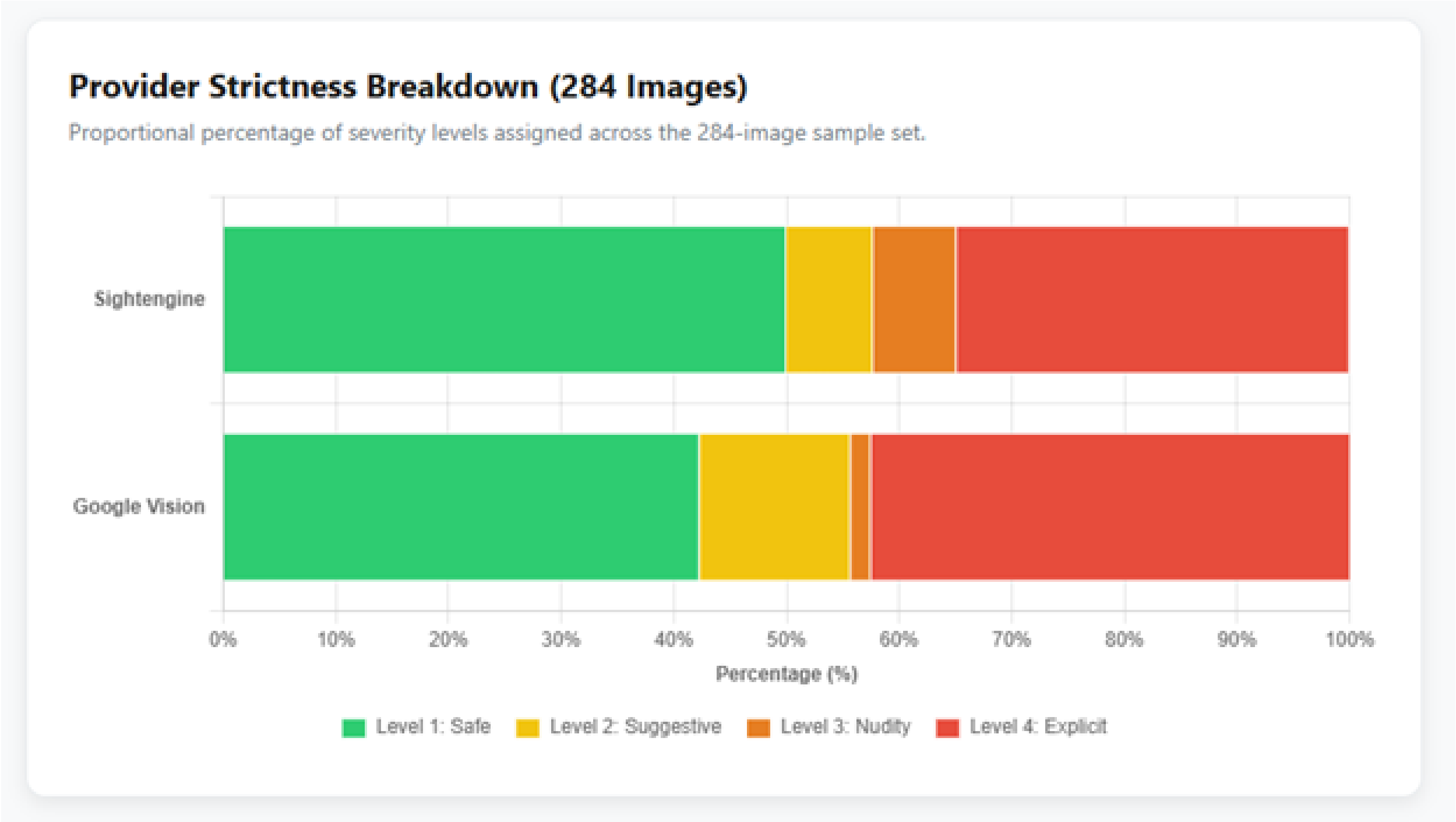

**Figure 3**. Distribution of Normalized Levels Across Sightengine and Google Vision

### 4.2. Scenario 2 - Detection and Classification of Person Age (Child vs Adult)

A critical component of CSAM detection is distinguishing minors from adults. Scenario 2 evaluated binary child-adult age classification using an age-labelled dataset and the minor-probability output of Sightengine.

The FG-NET Aging Database [5] was selected. It contains 1,002 images of 82 individuals aged 0-69 years, with multiple images per individual across different ages. The evaluation covered all 1,002 images in the dataset. Four images returned zero detected faces and did not produce a valid age-classification result. These were recorded as detection failures and excluded from the classification-performance calculations, leaving 998 valid classifications.

Images of individuals younger than 18 years were assigned the ground-truth class CHILD, while images of individuals aged 18 years or older were assigned ADULT. Sightengine's minor probability and its default classification decision were used for the evaluation. Performance metrics were calculated by comparing the returned predictions against the ground-truth age-derived classes.

| Metric | Value | Definition / Result |
|---|---|---|
| **Images in dataset** | **1002** | Total images in the FG-NET dataset |
| **Detection Failures** | **4** | Images with zero detected faces |
| **Images evaluated** | **998** | Total images in the FG-NET dataset |
| **Overall Accuracy** | **86.27%** | 861/998 correctly classified |
| **Precision (CHILD)** | **98.83%** | 506/512 predicted children were actual children |
| **Recall / Sensitivity (CHILD)** | **79.43%** | 506/637 actual children correctly identified |
| **Specificity (Adult)** | **98.34%** | 355/361 adults correctly identified |
| **F1-score (CHILD)** | **88.08%** | Harmonic mean of child precision and recall |

**Table 6**. Overall Child-Adult Classification Performance.

The results show high precision when Sightengine assigns the CHILD class, but lower recall. Of the 512 images classified as children, 506 were correctly classified, while six depicted adults. Conversely, 131 of the 637 child images were classified as adults. The system therefore missed a substantial proportion of minors, despite rarely classifying adults as children.

The corresponding confusion matrix was as follows:

| Ground Truth | Predicted CHILD | Predicted ADULT | Total |
|---|---|---|---|
| CHILD | 506 | 131 | 637 |
| ADULT | 6 | 355 | 361 |
| Total | 512 | 486 | 998 |

The principal observed error was the classification of minors as adults. Nevertheless, the six adult images classified as children demonstrate that adult specificity was high, but not perfect, in this evaluation.

### *Classification performance across age groups*

To investigate the relationship between chronological age and classification performance, the 998 valid classifications were grouped into five age brackets. Mean minor probability is reported on a scale from 0 to 1, with higher values indicating a stronger prediction of minority.

| Age Bracket | Sample Size (n) | Correct | Accuracy (%) | Mean Minor Prob. | Key Finding |
|---|---|---|---|---|---|
| **Early Childhood (0–5 yrs)** | 231 | 230 | **99.57%** | 0.975 | Near-perfect classification and very high mean minor probability |
| **Middle Childhood (6–12 yrs)** | 248 | 234 | **94.35%** | 0.902 | High accuracy, but some children were classified as adults |
| **Adolescence / Teens (13–17 yrs)** | 158 | 42 | **26.58%** | 0.292 | Major performance degradation, most adolescents were classified as adults |
| **Young Adults (18–30 yrs)** | 232 | 226 | **97.41%** | 0.051 | High accuracy and generally low minor probabilities |
| **Older Adults (31+ yrs)** | 129 | 129 | **100.0%** | 0.012 | All images correctly classified as adults, with consistently low minor probabilities |

**Table 7**. Classification Performance Across Age Groups.

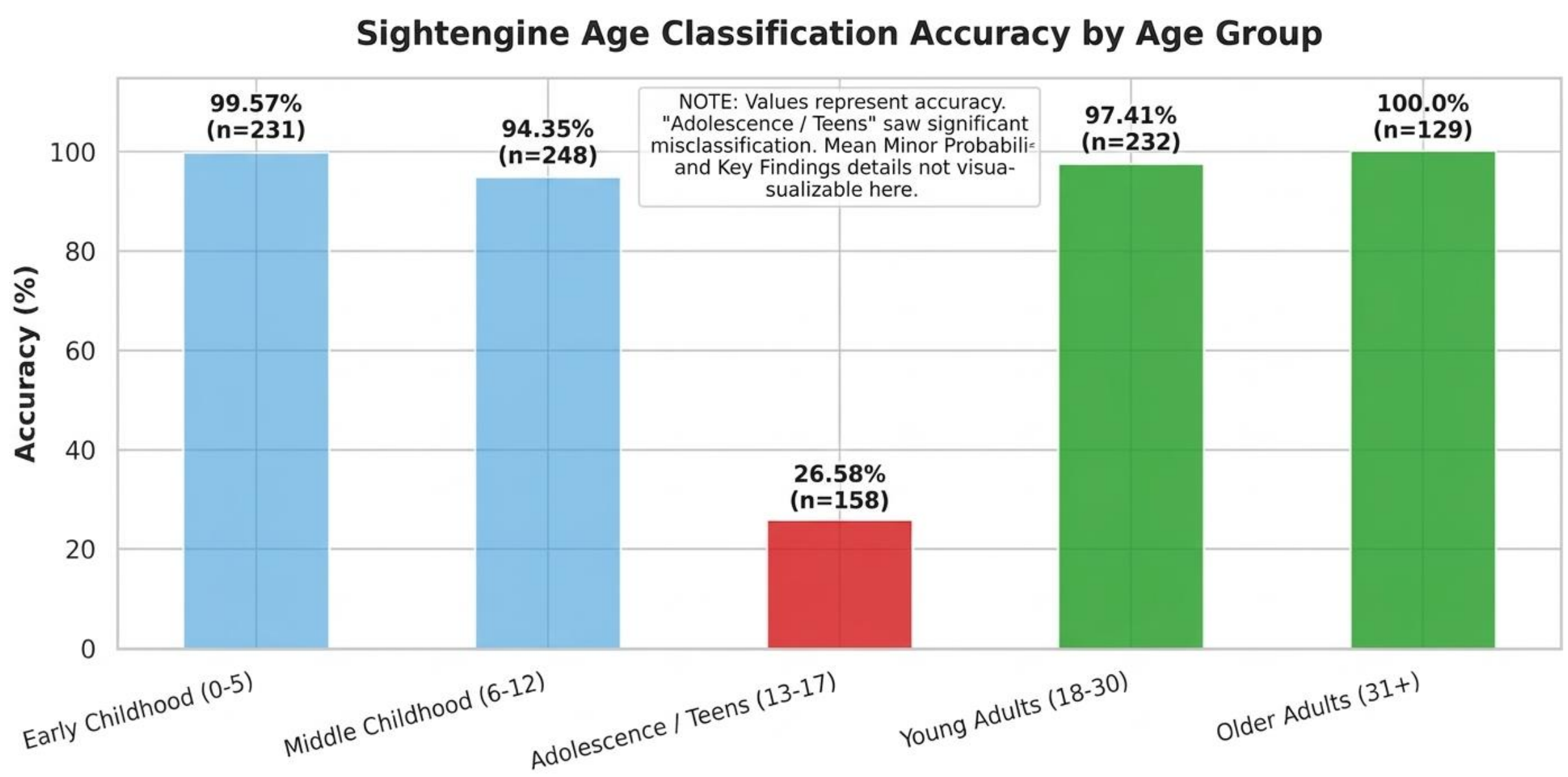

**Figure 4**. Age Classification Accuracy Across Age Groups.

The results reveal a pronounced age-dependent pattern in classification performance. Accuracy was particularly high for younger children, reaching 99.57% among those aged 0–5 and 94.35% among those aged 6–12. However, performance deteriorated sharply in the 13–17 age group, where only 42 of 158 images were correctly classified, corresponding to an accuracy of 26.58%. The remaining 116 images (73.42%) were classified as adults. In contrast, accuracy recovered substantially in the adult groups, reaching 97.41% among young adults and 100% among individuals aged 31 or older.

Mean minor probability followed a similar age-related pattern. It was 0.975 in early childhood and 0.902 in middle childhood, before dropping to 0.292 among adolescents. The corresponding values were considerably lower for young adults (0.051) and older adults (0.012). These results indicate a substantial reduction in the model's minor-probability scores during adolescence, although group-level means alone do not capture the full distribution or overlap of individual predictions.

The particularly low accuracy observed among adolescents highlights a limitation in the service's ability to distinguish older minors from adults. While Sightengine generally assigned high minor probabilities to younger children, this behavior did not extend consistently to individuals aged 13–17. This distinction is especially relevant for child-safety applications, as aggregate performance can conceal substantial differences between age groups.

The classification results also need to be considered in relation to the decision threshold. Under the service's default decision rule, the model achieved high CHILD precision (98.83%), but lower CHILD recall (79.43%). This indicates that, although predictions of CHILD were generally reliable, a considerable proportion of actual minors were not identified as such. Lowering the threshold for assigning the CHILD class could potentially recover some of these cases, but would also risk increase the number of adults incorrectly classified as children. The present experiment evaluates the service using its default decision rule and does not establish an optimal alternative threshold. Such an assessment would require examining the full distribution of minor probabilities across candidate thresholds and reporting threshold-specific confusion matrices, precision–recall curves and, where appropriate, ROC analysis. Any alternative decision threshold would need to account for the trade-off between missing minors and incorrectly classifying adults as children.

Overall, across the 998 valid classifications, Sightengine achieved an accuracy of 86.27%, with a CHILD precision of 98.83% and a CHILD recall of 79.43%. Of the 637 images depicting minors, 131 were classified as adults. Additionally, six adult images were incorrectly classified as children, resulting in an adult specificity of 98.34%. These figures demonstrate that, despite the high precision of CHILD predictions, the service's ability to identify minors is constrained by a substantial number of false negatives, particularly in the adolescent age group.

For CSAM/CSAE detection workflows, these findings support treating automated child-adult classification as an auxiliary indicator rather than a definitive age determination. The classification output could assist in prioritizing content for further examination, but an ADULT prediction should not be interpreted as reliable evidence that an individual is an adult. Similarly, a CHILD prediction does not constitute conclusive evidence of minority, given the observed false-positive classifications among adults. Particular caution is warranted when interpreting predictions involving adolescents, for whom the observed classification performance was considerably lower.

Finally, the results must be interpreted in light of the dataset's characteristics. Although all 1,002 images in the FG-NET Aging Database were analyzed, four images did not produce valid classifications, leaving 998 observations for the evaluation. Furthermore, the dataset contains images of only 82 individuals, with multiple images per person captured at different ages. Consequently, the 998 image classifications should not be considered 998 independent observations of different individuals. The findings therefore describe the service's performance on this particular dataset and should not be interpreted as evidence of universal performance or generalizability to other populations, image conditions or operational CSAM/CSAE datasets. Further evaluation using more diverse datasets and subject-independent testing would be necessary before drawing broader conclusions about the service's operational suitability.

### 4.3. Observations from the Practical Evaluation

The practical evaluation revealed several issues that are not apparent from provider specifications alone.

First, output taxonomies differ considerably. The sexual/nudity experiment demonstrated that apparently equivalent moderation functions can use substantially different terminology and levels of granularity. The normalized results showed 81.3% between Sightengine and Google Vision. This difference cannot be interpreted directly as difference in accuracy or strictness because it is affected by provider-specific taxonomies and the normalization procedure. Retaining the raw vendor outputs is therefore essential.

Second, normalization facilitates comparison but introduces methodological assumptions. Mapping different classifications to a common four-level taxonomy makes high-level comparison possible, but it may obscure distinctions between the underlying models. This is particularly relevant to Google SafeSearch, which does not explicitly distinguish all the categories represented in the taxonomy. Normalized classifications should therefore be interpreted alongside the original vendor outputs rather than as directly interchangeable measurements.

Third, age-classification performance varies substantially with age. The Sightengine experiment evaluated the full set of 1,002 FG-NET images, of which 998 produced valid classifications. Overall accuracy was 86.27%, but this aggregate result concealed a pronounced weakness in identifying adolescents. Accuracy was 99.57% for children aged 0-5 and 94.35% for those aged 6-12, but fell to 26.58% for the 13-17 age group. Of the 158 valid adolescent images, 116 were classified as adults. Overall child recall was 79.43%, while CHILD precision was 98.83% and adult specificity was 98.34%. These results demonstrate that strong performance on younger children can conceal a critical weakness in identifying older minors.

Fourth, confidence and probability outputs require careful interpretation. The tested services produce assessments whose meaning and scale depend on the provider and the task. In age classification, mean minor probability decreased from 0.975 in early childhood and 0.902 in middle childhood to 0.292 among adolescents. This pattern highlights the potential importance of decision-threshold selection. Lowering the threshold for assigning the CHILD class could recover some minors currently classified as adults, but could also increase the number of adults incorrectly classified as children. A formal threshold analysis using the full probability distribution and threshold-specific performance metrics would be necessary before recommending an alternative threshold.

Fifth, dataset composition and experimental scope limit generalization. The sexual/nudity experiment used a sample of 288 images, with 284 available for direct two-platform comparison, while the age-classification experiment covered all 1,002 FG-NET images, of which 998 produced valid classifications. However, FG-NET contains images of only 82 individuals, with multiple images per individual at different ages. The results should therefore not automatically be interpreted as evidence of generalization to previously unseen individuals. Neither experiment establishes universal performance across all image types, demographic groups, or operational environments. Larger and more diverse datasets, together with subject-independent evaluations, would be needed for stronger operational conclusions.

Finally, the experiments reinforce the conclusions of the preceding technical analysis: AI-based image analysis should be regarded as an auxiliary decision-support mechanism, not an autonomous means of determining whether material constitutes CSAM. The evaluated services demonstrated useful capabilities for classifying content and generating age-related indicators, but differences in classification behavior, output taxonomies, probability thresholds, and missing results require careful interpretation and human review. In particular, an ADULT prediction should not be treated as reliable evidence that a person is an adult, given the substantial number of adolescent images classified as adults.

Overall, practical usefulness depends not only on advertised AI capabilities but also on output structure, classification granularity, confidence information, compatibility with reference data, dataset representativeness, and the ability to integrate heterogeneous outputs into a common workflow. These factors should therefore be considered when evaluating technologies for potential integration into the hash-checking ecosystem such as the PreventCSA@EU system.

## 5. Legal Dimensions of AI-Based CSAM Detection

The technologies described above are increasingly deployed by online service providers as part of broader content moderation and detection frameworks, including for the identification and prioritization of content potentially constituting CSAM. Their growing integration into private-sector content moderation practices raises important questions concerning the legal framework governing their deployment.

### 5.1. Regulatory Asymmetries in AI-Based CSAM Detection

Although the European Data Protection Board (EDPB) [30], when assessing the initial draft of the CSAR Proposal [31], regarded AI-based CSAM detection tools as highly intrusive given their implications for privacy and the confidentiality of communications, the AI Act [32] does not classify such systems, when deployed by private service providers, as high-risk solely on the basis of the sensitive nature of the content they process. Consequently, notwithstanding their potential to significantly affect fundamental rights, their deployment by private service providers does not, on that basis alone, trigger the enhanced requirements associated with the AI Act's high-risk regime. By contrast, where AI-based detection tools are deployed by law enforcement authorities (LEAs), they may fall within the category of high-risk AI systems pursuant to Article 6(2) of the AI Act where the intended use falls within one of the law-enforcement use cases listed in Annex III. In such cases, the deployment of these systems may trigger the requirement to conduct a Fundamental Rights Impact Assessment (FRIA) under Article 27.

In particular, Annex III (6) [32] classifies as high-risk AI systems intended to be used by, or on behalf of, law enforcement authorities, or by Union institutions, bodies or agencies in support of law enforcement authorities, for evaluating the reliability of evidence during the investigation or prosecution of criminal offences. The same classification applies to AI systems used by LEAs, or on their behalf, for assessing the risk that a natural person may offend or re-offend, provided that such assessments are not based solely on the profiling of natural persons[3].

It should be noted, however, that the use of AI-based CSAM detection technologies by LEAs does not automatically bring such systems within the scope of Annex III (6) of the AI Act. Their classification depends on the intended purpose and the specific law-enforcement function for which they are deployed, as Annex III (6) identifies specific functions, such as the evaluation of the reliability of evidence, rather than categories of criminal offences. Accordingly, the use of an AI system merely to identify or classify visual content as potentially constituting CSAM would not, in itself, fall within Annex III (6).

[3] Profiling of natural persons within the meaning of Article 3(4) of Directive (EU) 2016/680.

Alongside the AI Act, AI-based CSAM detection by private service providers is also governed by an evolving EU framework for the detection and handling of CSAM on their services. Successive legislative changes and temporary arrangements have prevented this framework from crystallizing into a stable regulatory regime. The proposed Child Sexual Abuse Regulation has not yet been adopted, while the derogation established by Regulation (EU) 2021/1232 [33] expired on 3 April 2026, leaving no specific EU framework providing for such a derogation until Regulation (EU) 2026/1881 [34] entered into force on 31 July 2026. During this intervening period, several private service providers, including Google, Meta, Microsoft and Snap, continued to implement voluntary measures to combat online child sexual abuse under their respective community policies [35].

Regulation (EU) 2026/1881 has since re-established a temporary derogation from Articles 5(1) and 6(1) of the e-Privacy Directive, subject to specific safeguards. In particular, Article 3(1)(c) requires a prior DPIA under Article 35 GDPR [36] and prior consultation under Article 36 GDPR in respect of any specific technology used to detect and remove online CSAM and report it.

Against this regulatory background, functionally comparable AI-based CSAM detection systems may, in the event of the establishment of the EU Centre envisaged by the CSAR Proposal [32][4], operate within an increasingly interconnected detection, reporting and investigative ecosystem while remaining subject to different ex ante assessment requirements. This differentiation depends on the institutional and functional context in which the system is deployed, in particular on whether it is used by a private service provider or a law enforcement authority and, in the latter case, on the specific law-enforcement function for which it is employed. Thus, while private-provider deployment is subject to the safeguards described above, the deployment of a functionally comparable system by a law enforcement authority may, where it falls within a high-risk use case under Annex III (6) of the AI Act, additionally trigger the obligation to conduct a FRIA under Article 27, alongside any applicable DPIA under Article 27 of the Law Enforcement Directive.

The interconnection between these different contexts can be illustrated through the pathways by which material, indicators and technology-generated outputs may be used across the envisaged ecosystem. A private service provider may use detection technology to identify suspected CSAM and submit a corresponding report to the EU Centre, which may subsequently forward the report to the competent national law enforcement authorities for further assessment and investigation [32] [37][5]. Material identified by the competent national authorities as constituting CSAM may, through the procedure envisaged by the CSAR Proposal, be submitted to the EU Centre and form the basis for the generation of indicators maintained in its databases [38][6]. Within the envisaged CSAR architecture, these indicators may subsequently support detection by private service providers through the detection infrastructure and technologies made available to them by the EU Centre [39][7]. At the same time, the EU Centre's indicator databases may be accessed by competent national LEAs, to the extent necessary for the investigation of suspected child sexual abuse offences [40][8].

[4] European Commission, Proposal for a Regulation laying down rules to prevent and combat child sexual abuse, COM (2022) 209 final; see also the subsequent Council Presidency compromise texts and documents concerning the ongoing interinstitutional negotiations, including Council documents 6312/26, 9659/26 and 10431/26. The analysis in this section takes account of the state of the interinstitutional negotiations under the Cyprus Presidency as reflected in the relevant Presidency documents available at the time of writing.

[5] European Commission, Proposal for a Regulation laying down rules to prevent and combat child sexual abuse, COM (2022) 209 final; see also the subsequent Council Presidency compromise texts and documents concerning the ongoing interinstitutional negotiations, including Council documents 6312/26, 9659/26 and 10431/26. The analysis in this section takes account of the state of the interinstitutional negotiations under the Cyprus Presidency as reflected in the relevant Presidency documents available at the time of writing.

[6] European Commission, Proposal for a Regulation of the European Parliament and of the Council laying down rules to prevent and combat child sexual abuse, COM (2022) 209 final, Arts 36(1)(a) and 44(1)-(3). See also Council of the European Union, Presidency

Beyond this pathway, law enforcement authorities may deploy AI-based technologies in criminal investigations without the outputs or indicators generated or used necessarily being incorporated into the EU Centre's databases, with such deployments additionally triggering the FRIA requirement under Article 27 where they fall within a high-risk use case under Annex III (6) of the AI Act [32]. The resulting architecture is therefore not merely a linear reporting chain, but a network of interconnected yet institutionally and functionally distinct technological deployments. This interconnected architecture is particularly relevant to the ex-ante assessment of the fundamental-rights implications arising across these different deployment contexts, including for the rights and protection of children depicted in abusive material.

Although functionally comparable technologies may be used across different stages of the detection, reporting and investigative process, their ex ante fundamental-rights assessment remains dependent on the identity of the actor, the relevant processing operation and the specific deployment context, potentially resulting in fragmentation across distinct, albeit partly overlapping, legal frameworks. This differentiation is particularly significant from a children's-rights perspective: while certain law-enforcement deployments may trigger a FRIA specifically directed at the fundamental-rights implications of the deployment, no equivalent FRIA requirement applies to private-provider CSAM detection as such. This raises the question whether the applicable assessment framework adequately captures the rights and interests of children affected by these systems, particularly where detection, reporting and investigative activities form part of the same broader process.

While both DPIAs and FRIAs follow a risk-based approach and seek to protect the rights and freedoms of affected individuals, a DPIA is centered on the risks arising from envisaged personal-data processing operations, although the "rights and freedoms" considered may extend beyond privacy and data protection to other fundamental rights [41]. By contrast, the FRIA under Article 27 of the AI Act is linked to the concrete deployment of certain high-risk AI systems and extends to the fundamental-rights risks arising in their specific context of use [42].

The two assessments may therefore overlap but are not interchangeable [32]. Their complementarity is expressly recognized by Article 27(4) of the AI Act, which allows relevant obligations already addressed through a DPIA under the GDPR or the Law Enforcement Directive to be considered within the FRIA. The FRIA thus complements rather than replaces any applicable DPIA. This differentiated assessment framework is particularly significant in the context of AI-based CSAM detection. Children depicted in

---

compromise texts on detection, 9659/26, 25 May 2026, referring to indicators from the databases operated by the EU Centre pursuant to Article 44.

[7] Council of the European Union, Presidency compromise texts on detection, Council doc 9659/26, 25 May 2026, in particular the proposed provisions on own-initiative searches, requiring providers to use the corresponding indicators provided by the EU Centre in accordance with Article 44; Council of the European Union, State of play of interinstitutional negotiations and Presidency compromise texts, Council document 10431/26, 12 June 2026, proposed Article 43(e), concerning the EU Centre's task of making technologies available to providers of relevant information society services for the identification of online child sexual abuse.

[8] European Commission, Proposal for a Regulation of the European Parliament and of the Council laying down rules to prevent and combat child sexual abuse, COM (2022) 209 final, Art. 46(4); see also Council of the European Union, WK 3494/2026 INIT, Four-column table with the outcome of the trilogue on 26 February 2026, 6 March 2026, Art. 46(4).

abusive material are not merely data subjects but rights-holders whose rights, dignity and protection may be affected by the operation and effectiveness of the system.

Article 27 of the AI Act requires consideration of the persons and groups likely to be affected and of the specific risks of harm arising from deployment, providing a framework within which impacts on children and other affected groups may be considered, including in light of Article 24 of the Charter and the best interests of the child. In this context, false positives and false negatives illustrate how fundamental-rights risks may arise in different ways and affect different groups [43]. False positives may interfere with the rights of users and other affected persons, whereas false negatives may leave actual CSAM undetected, allowing its continued circulation and potentially perpetuating harm to the rights, dignity and protection of the children depicted.

The differentiated framework for assessing fundamental-rights impacts becomes particularly significant in light of broader concerns raised in the literature regarding the evolving role of private actors within this ecosystem. While private providers do not formally exercise State authority, they are increasingly involved in activities that intersect with functions traditionally associated with public law enforcement, particularly where the detection of potentially illegal content leads to reporting to competent authorities and subsequent criminal investigations [44]. This evolving role further underscores the importance of ensuring that differences in the applicable ex ante assessment frameworks do not translate into uneven levels of fundamental-rights protection across the detection, reporting and investigative process.

### 5.2 Algorithmic Inferences and Legal Implications

The inference stage constitutes a critical component of the logic underpinning AI systems. It is at this stage that raw data is transformed into predictions, classifications, or assessments concerning individuals, often generating new information that was not explicitly provided by the data subject. The significance of such derived information has long been recognized in data protection law. In its Guidelines on Automated Individual Decision-Making and Profiling, the Article 29 Working Party observed that profiling may involve statistical deductions through which information about an individual is inferred from data originating from different sources [45].

The legal significance of algorithmically generated outputs is illustrated by the CJEU's judgment in SCHUFA Holding (Case C-634/21), where the Court held that the automated generation of a credit-scoring probability value may itself constitute automated individual decision-making within the meaning of Article 22(1) GDPR where a third party, to which that value is transmitted, draws strongly on it in establishing, implementing or terminating a contractual relationship with the data subject [46]. An algorithmic output cannot therefore necessarily be regarded as a merely preparatory act: where it plays a determining role in the action taken by a third party and thereby significantly affects an individual, its generation may itself fall within the concept of a "decision" under Article 22(1). In this context, Articles 13(2)(f), 14(2)(g) and 15(1)(h) GDPR require [36], in the circumstances specified therein, meaningful information about the logic involved, as well as the significance and envisaged consequences of such processing, while Article 22 provides safeguards in relation to solely automated decisions producing legal or similarly significant effects.

Although the extent to which these provisions entail an individualized explanation has been debated [47], the CJEU clarified in Dun & Bradstreet Austria (Case C-203/22) that Article 15(1)(h) GDPR affords data subjects a genuine right to an explanation of the automated decision-making concerned and its result. "Meaningful information about the logic involved" must therefore explain the procedure and principles actually applied to the data subject's personal data in reaching the specific result, in a manner enabling the data subject to understand which personal data were used and how they were used in the automated decision-making.

The AI Act complements this framework through transparency requirements for high-risk AI systems, human-oversight requirements, and corresponding obligations on deployers under Articles 13, 14 and 26 [32]. Most significantly, Article 86 establishes an explicit but limited right to explanation for persons affected by decisions taken by deployers on the basis of the output of certain high-risk AI systems listed in Annex III [50]. Where such a decision produces legal effects or similarly significantly affects the person in a way that they consider to have an adverse impact on their health, safety or fundamental rights, they are entitled to a clear and meaningful explanation of the role of the AI system in the decision-making procedure and the main elements of the decision taken [32].

Its relevance to AI-based CSAM detection is, however, constrained by this limited scope. As discussed in the preceding section, AI-based CSAM detection systems deployed by private service providers do not automatically fall within the high-risk categories of Annex III. Consequently, even where an AI-generated inference contributes to a decision significantly affecting a user, the specific right to explanation under Article 86 may not apply. This reflects a broader limitation identified in the literature: restricting AI Act explanation rights to high-risk applications may leave protection gaps where AI systems fall outside that classification while still affecting individuals [48].

This limitation becomes particularly significant where private service providers use AI-based behavioral analysis to identify patterns suggesting that a user may be disseminating CSAM. Such inferences may contribute to subsequent measures, including the removal of content, reporting to competent hotlines, the National Center for Missing & Exploited Children (NCMEC) or law enforcement authorities, and the suspension or termination of user accounts. Depending on the circumstances, these measures may have significant consequences for the individuals concerned, including implications for privacy, data protection, freedom of expression and access to digital services [49]. Yet the seriousness of these downstream consequences does not, by itself, bring the underlying AI system within the scope of Article 86.

This does not necessarily leave affected users without explanatory safeguards. Where a hosting service removes content or suspends or terminates an account on the ground that information provided by the user is illegal or incompatible with its terms and conditions, Article 17 DSA requires a clear and specific statement of reasons, including, where applicable, information on the use of automated means in taking the decision [50].

The literature has similarly noted that AI-based content moderation may give rise to explanation obligations under Article 17 DSA while remaining outside Article 86 AI Act, since content moderation is not itself listed as a high-risk application in Annex III [48]. The two provisions, however, structure their explanatory obligations differently. While Article 17 DSA requires information concerning the moderation measure, the relevant factual circumstances and legal or contractual grounds and, where applicable, the use of automated means, Article 86 requires a clear and meaningful explanation of the role of the AI system in the decision-making procedure and the main elements of the decision taken. The explanatory safeguards available to affected users may therefore differ in their scope and content depending on the regulatory framework applicable to the particular decision.

## 6. Conclusions

From a technical perspective, the analysis and practical evaluation demonstrate that AI-based image-analysis and content-moderation technologies can complement hash-based detection by providing additional signals for previously unknown material, including classification and moderation labels, confidence or risk scores, face and object detections, age-related estimates, and semantic metadata. Their practical value, however, depends not only on model capabilities but also on API accessibility, output granularity, confidence information, the alignment between the evaluated task and the reference dataset, and the interoperability of heterogeneous provider outputs.

The experiments further showed that provider-specific taxonomies and thresholds can materially affect cross-platform comparisons, while age-related classification performance may vary substantially across age groups, particularly near the child–adult boundary. These findings support the use of AI-based analysis as a complementary triage and decision-support mechanism rather than as an autonomous means of determining whether material constitutes CSAM. Within the hash checking ecosystem (such as the PreventCSA@EU system), such tools could therefore extend existing hash-based detection capabilities for previously unknown material, provided that their outputs are interpreted in context and remain subject to appropriate human review.

From a legal perspective, the analysis demonstrates that the regulatory treatment of AI-based CSAM detection in the EU remains highly context-dependent. The sensitive nature of the technology does not, in itself, determine its classification under the AI Act. Rather, the deploying actor, intended purpose and concrete use determine the applicable obligations. Functionally comparable technologies may therefore be subject to different ex ante safeguards: while private-provider deployment is subject to applicable data-protection requirements and the safeguards established by Regulation (EU) 2026/1881, certain law-enforcement deployments may additionally trigger a FRIA under Article 27 of the AI Act, alongside an applicable DPIA under the Law Enforcement Directive.

This differentiation becomes particularly significant within an interconnected detection, reporting and investigative ecosystem, where technologically generated outputs and indicators may move across private providers, the envisaged EU Centre and law enforcement authorities, while fundamental-rights assessment remains tied to individual deployment contexts. This fragmentation may prevent a comprehensive assessment of risks arising across the broader detection process, including the risk that false negatives allow actual CSAM to remain undetected and in circulation, thereby perpetuating harm to the rights, dignity and protection of the children depicted.

A similar differentiation arises in relation to explainability. AI-generated inferences may contribute to measures significantly affecting users without necessarily triggering the specific right to explanation under Article 86 AI Act. Although Article 17 DSA may provide explanatory safeguards for certain content-moderation decisions, these are not equivalent in scope and content to the explanation provided for under Article 86. Taken together, these findings underline the need for proportionate and coherent safeguards capable of addressing fundamental-rights risks across the interconnected detection, reporting and investigative ecosystem, while considering the distinct roles and functions of private providers and law enforcement authorities.

## Acknowledgments

This work was supported by the Internal Security Fund of the European Union under Specific Action "Towards a Coordinated and Cooperative Effort for the Prevention of Child Sexual Abuse at a European level" (Call: ISF/2022/SA/1.4.1; MIS: 6002910). The views expressed are those of the authors and do not necessarily reflect the view of the European Commission.